\documentclass[10pt,twocolumn,letterpaper]{article}

\usepackage[pagenumbers]{cvpr} 
\usepackage[accsupp]{axessibility}  

\usepackage{graphicx}
\usepackage{soul}
\usepackage{amsmath, mathtools}
\usepackage{amssymb}
\usepackage{booktabs}
\usepackage{algorithmic}
\usepackage{algorithm}
\newtheorem{definition}{Definition}[section]
\newtheorem{theorem}{Theorem}[section]
\usepackage{multirow}
\usepackage[pagebackref,breaklinks,colorlinks]{hyperref}

\usepackage[capitalize]{cleveref}
\crefname{section}{Sec.}{Secs.}
\Crefname{section}{Section}{Sections}
\Crefname{table}{Table}{Tables}
\crefname{table}{Tab.}{Tabs.}

\def\cvprPaperID{*****} 
\def\confName{CVPR}
\def\confYear{2023}

\begin{document}

\title{Hamming Similarity and Graph Laplacians for Class Partitioning and Adversarial Image Detection}
\author{Huma Jamil$^{1}$\thanks{Equal contributors.},
        Yajing Liu$^{2}$\footnotemark[1] \thanks{Corresponding author.},
        Turgay Caglar$^{1}$,
        Christina M. Cole$^{2}$,
        Nathaniel Blanchard$^{1}$,\\
        Christopher Peterson$^{2}$,
        Michael Kirby$^{2}$\\
$^1$Computer Science Department, Colorado State University, Fort Collins, CO, USA\\
$^2$Mathematics Department, Colorado State University, Fort Collins, CO, USA\\
Emails: \emph{\{huma97, yajing.liu, turguy.caglar, christina.rigsby, nblancha\}@colostate.edu}, \\
\emph{\{christopher2.peterson, michael.kirby\}@colostate.edu}
}

\maketitle

\begin{abstract}

Researchers typically investigate neural network representations by examining activation outputs for one or more layers of a network. Here, we investigate the potential for ReLU activation patterns (encoded as bit vectors) to aid in understanding and interpreting the behavior of neural networks. We utilize Representational Dissimilarity Matrices (RDMs) to investigate the coherence of data within the embedding spaces of a deep neural network. From each layer of a network, we extract and utilize bit vectors to construct similarity scores between images. From these similarity scores, we build a similarity matrix for a collection of images drawn from 2 classes. We then apply Fiedler partitioning to the associated Laplacian matrix to separate the classes. Our results indicate, through bit vector representations, that the network continues to refine class detectability with the last ReLU layer achieving better than 95\% separation accuracy. Additionally, we demonstrate that bit vectors aid in adversarial image detection, again achieving over 95\% accuracy in separating adversarial and non-adversarial images using a simple classifier. 

\end{abstract}

\section{Introduction}
\label{sec:intro}
For nearly as long as neural networks have dominated various benchmarks, works have investigated methods for comprehending and explaining their behavior. The recent controversies around when to, and not to, trust the output from ChatGPT perfectly encapsulates the problems with treating networks as a ``black box" \cite{allyn_microsofts_2023} --- explaining network behavior links to trust, and the potential to better understand how to enhance performance even further. One method to accomplish this is by examining the coherence of input data within the lower-dimensional embedding spaces that exist within the network, as advocated by the seminal work on representation learning \cite{bengio2013representation}. Ideally, these intermediate embeddings will represent data in such a manner that prescribed similarities (i.e. same class label) can be identified and distinguished from prescribed dissimilarities (i.e. distinct class labels). 
Geometrically, this manifests if data identified as similar are mapped to points that are close together in some embedding space while data identified as dissimilar are mapped to points that are further apart. It is then a simple task to distinguish similar data from dissimilar data. 

Analyzing how these embedding spaces are utilized and their effectiveness can give insights into how the network is processing and representing the data as it passes through. This in turn can be used to improve the performance of the neural network and facilitate the interpretation of its results. 
Representational dissimilarity matrices (RDMs) are a useful tool for visualizing the degree of coherence within these embedding spaces --- RDMs quantify the (dis)similarity of a network's representation to like- and dislike stimuli. By examining the RDMs, we can identify patterns or irregularities in the network's representations of input data in order to diagnose and address problems with the model's performance or interpretability \cite{kriegeskorte2008representational}. We extend these RDMs with a novel analysis: extending Fiedler partitioning to dissimilarity matrices interpreted as weighted graphs. Fiedler partitioning \cite{Fiedler1973, Fiedler1975}, also known as spectral partitioning, is a technique used in graph theory to divide a graph into two or more subgraphs based on the eigenvalues and eigenvectors of the graph's Laplacian matrix. 

Typically, the activations for a layer of a neural network are utilized in neural network pattern analysis to unravel and encapsulate useful statistics about the input \cite{olah2017feature}. However, recent research has demonstrated that the coarser utilization of ``bit vectors", derived from the output of Rectified Linear Unit (ReLU) activation patterns in early layers, are meaningfully linked to adversarial images~\cite{jamil2022dual}, which inspired us to use RDMs to further investigate the representations of bit vectors. 
In our study, we investigated the effectiveness of Fiedler partitioning as derived from the Laplacian of a similarity matrix constructed from the bit vectors within a layer of a neural network. In other words, we computed the Fiedler eigenvector of a certain Laplacian matrix related to the pattern of ReLU activations in the nodes of a layer of the network. Specifically, we examined the RDMs of each ReLU layer for two classes using the Fiedler partitioning. Since the number of nodes in a layer of a deep neural networks can be quite large (ranging from 100k to 800k in our case), we hypothesized that a large number of bits were not really needed for the classification task.
To illustrate this, we identified and extracted 1000 of the most significant bits that ultimately led to the construction of the RDMs.

Our results revealed that as data proceeded to deeper ReLU layers, distinct partitions began to emerge in the data, and the classification accuracy for each class generally improved. An important observation from our study is that layers did not treat all data equally; some layers appeared to be specifically focused on improving the ability to distinguish between a smaller collection of images.
The RDM analysis offers insights into how a network encodes information about the input data and extracts crucial features for classification, improving its transparency and interpretability.

Based on the above analysis, we then conducted empirical experiments to evaluate the effectiveness of bit vector-induced partitioning to distinguish adversarial images from non-adversarial images. Previous work has shown that adversarial and non-adversarial images are linearly separable in the latent layer {(the penultimate layer activation)} embeddings of a deep neural network \cite{Gorbett_2022_WACV}. Although bit vectors provide a more coarse level of information about input data, we have found that bit vectors from the last ReLU layer of a neural network can also be used to effectively separate non-adversarial images from adversarial ones. In fact, we produced results that were at least as good as those achieved by latent layer embeddings while using fewer features.
Overall, our results suggest that bit vectors in deep neural networks offer a promising approach for identifying adversarial images  with high efficiency and potentially improved performance.

Our contributions in this paper are as follows:
\begin{itemize}
    \item We demonstrate that bit vector subsampling utilizing 1000 out of the 100k-800k available bits can separate two-class data.
    \item We present a novel idea of utilizing Fiedler partitioning on the Laplacian matrix constructed from bit vector measures of similarity to achieve over 95\% image class identification accuracy.
    \item We show that bit vector representations can separate adversarial images from non-adversarial ones with more than 95\% accuracy using a linear Support Vector Machine (SVM) .

\end{itemize}

The structure of our paper is as follows: Section~\ref{sec:relatedwork} provides a comprehensive review of the related literature, including Fiedler partitioning, adversarial detection, and network interpretability. Section~\ref{sec:definitions} outlines the relevant definitions that underpin the study. Section~\ref{sec:Fiedler} presents the Fiedler Vector algorithm and its practical application to the two-class and two-superclass classification of images via the bit vectors of ResNet-50. Section~\ref{sec:adversarial} illustrates that the linear separability of the bit vectors of ResNet-50 can be utilized to differentiate between non-adversarial and adversarial images. Finally, Section~\ref{sec:conclusion} offers concluding remarks.
\section {Related Work}
\label{sec:relatedwork}
\subsection{Fiedler Partitioning}
The Fiedler vector, an eigenvector corresponding to the smallest non-zero eigenvalue of the Laplacian matrix of a weighted graph, is a fundamental tool in graph theory for elucidating underlying substructures within a complex system represented through pairwise similarities. Its entries can be utilized to assign the vertices of the graph to two disjoint subsets while minimizing the sum of weights of the edges between the subsets and maximizing the total weight of edges inside the two subsets. The Fiedler vector algorithm, also called Fiedler partitioning, has been successfully applied to numerous real-world problems:
identifying regions with similar texture or color properties \cite{Shi2000}, identifying groups that share similar characteristics or functions \cite{PhysRevE.69.026113}, and highlighting influential nodes or communities in a social network \cite{Leskovec2008}.
It is also a well-established technique in machine learning and data analysis to uncover and interpret subpatterns within data \cite{duan2019improving, xie2016unsupervised}. One of the key advantages of the Fiedler partitioning is its ability to leverage a lower-dimensional feature space derived from the network, as demonstrated in recent research by Schleider et al. \cite{schleider2022study} and Cao et al. \cite{cao2022unsupervised}. This feature space, typically obtained via eigenvalue decomposition of a graph Laplacian matrix, allows for efficient and effective data clustering and visualization, making it a popular choice for a variety of applications. In this work, we apply Fiedler partitioning to representational dissimilarity matrices to perform novel analyses. 

\subsection{Adversarial Detection}
Deep neural networks are efficient in image classification, but can be tricked by adversarial attacks \cite{szegedy2013intriguing}. Several distinct attacks have been proposed to exploit various weaknesses \cite{mahloujifar2019curse,9238430,goodfellow2014explaining,madry2017towards,7958570}.
In response, multiple studies have been conducted to leverage the characteristics of the hidden and latent layers of deep neural networks for the purpose of detecting adversarial images.
For example, \cite{Bendale_2016_CVPR} proposed a method using the latent layers, and \cite{Gorbett_2022_WACV} showed the meaningful information carried by activations within these layers. 
To detect adversarial and out-of-distribution images, \cite{lee2018simple} used features from hidden and latent layers to compute class conditional Gaussian distributions and Mahalanobis distances. On the other hand, \cite{Li_2017_ICCV}  utilized intermediate convolution layer outputs to extract statistics that aid in detecting adversarial images. Our methodology bears similarities to \cite{Li_2017_ICCV}  in that we also examine the internal workings of the neural network. However, we employ a more rudimentary metric, specifically the bit vectors generated by a relatively small subset of a ReLU layer's activation pattern rather than relying on the latent layer values.
\subsection{Network Interpretability}
Interpreting the function of deep neural networks often involves understanding the hidden layers. Visualization techniques using intermediate layers \cite{10.1007/978-3-319-10590-1_53}, network dissection \cite{8417924}, and learning disentangled representations \cite{paige2017learning} are some of the few methods proposed to achieve this.  In our work, we contribute to network interpretability by using bit vectors extracted from intermediate ReLU layers.
In previous literature,  quantitative metrics like Hamming distance \cite{shamir2019simple, siddiqui2022trust}, Jaccard distance \cite{amparore2021trust},  and cosine similarity \cite{kim2018interpretability} have been used to evaluate the effectiveness of the explainable methods. We computed image similarity in our work by evaluating the pairwise Hamming distance between the bit vectors determined by data passing through a ReLU layer and constructing a Laplacian matrix based on these distances. Our results provide insights into the neural network's information processing, feature extraction, and decision-making mechanisms, enhancing the network's transparency and interpretability.

\section{ Definitions}
\label{sec:definitions}
\subsection{Bit Vectors}
Consider a feed forward neural network containing ReLU activation functions. Suppose the $i^{th}$ layer of the network contains $h_i$ ReLU nodes. For a given input $x\in\mathbb{R}^m$, we denote its output in the $i^{th}$ layer, at these ReLU nodes, as $o^i(x)=\begin{bmatrix} o^i_1(x)\ \ldots \ o^i_{h_i}(x) \end{bmatrix}^{\top}$. We define the bit vector of $x$ in the $i^{th}$ layer as $s^i(x) =\begin{bmatrix} s^i_1(x)\ \ldots \ s^i_{h_i}(x)\end{bmatrix}^{\top}$ with 
\begin{equation}
\label{eq:bitvec}
    s_j^i(x)\coloneqq \begin{cases}1 \ \quad\text{if}\ o_j^i(x)>0,\\
    0 \quad \ \text{if}\ o_j^i(x)=0.
    \end{cases}
\end{equation}
Thus, the bit vector is a function of the input $x$ and assigns a value of $1$ to nodes that activate the ReLU function at $x$ and assigns a value of $0$ otherwise. It is typical that the bit vectors for various inputs in the neural network will exhibit diverse activation patterns in the ReLU layers as the inputs traverse the network.

\subsection{Representational Dissimilarity Matrices}
For a given neural network $N$ with input set $X=\{x_1,\ldots, x_n\}$,  let $\left \{o^{i}(x_1), \ldots, o^i(x_n)  \right \}$ be the output of $X$ in the $i^{th}$ hidden layer of $N$, we define a representational dissimilarity matrix (RDM) at layer $i$ as:
\begin{equation}
\label{eq:rdm}
 RDM\left ( X,N \right )= \left ( d^i_{jk} \right )_{0\leq j,k\leq n}, \\
\end{equation}
where $d^i_{jk}$ is a measure of dissimilarity between $o^i(x_j)$ and $o^i(x_k)$. For instance, $d^i$ could be a distance function.
In essence, Representational Dissimilarity Matrices (RDMs) provide a way to capture the internal representations of a neural network by summarizing the similarities and differences between the neural responses to different inputs.

To define the RDM, one needs a measure of dissimilarity. Different measures lead to different RDMs. In our study, we employed the Hamming distance and the cosine distance for $d^i$. The Hamming distance is a metric commonly used for comparing two binary strings of the same length and denotes the number of positions where the corresponding entries are distinct. We utilize the Hamming distance as one of the distance metrics when constructing the RDM for an input set $X$ at layer $i$. The cosine distance between two vectors is defined as $1-\cos(\theta)$ where $\theta$ denotes the angle between the two vectors. In addition to the Hamming distance, we utilize the cosine distance as a dissimilarity score for the embeddings occuring in the latent layer of the neural network.


\subsection{Feature Selection}
\label{sec:featureselection}
Feature selection is a technique used to reduce the dimensionality of the data by choosing features that do a good job representing the data. In our experiments, we used \emph{SelectKBest} method from a Python library to reduce the dimensionality of bit vectors (i.e. to select a subset of bits from the bit vector). Mathematically, let $X=\{x_1,\ldots, x_n\}$ be a set of $n$ input data points and let $Y=\{y_1,\ldots, y_n\}$ be the target labels. Let $f_i(X,Y)$ be a scoring function that measures the importance of the $i^{th}$ feature in predicting the target variable. Then, the \emph{SelectKBest} function selects the $k$ features with the highest scores, and it returns the set of indices $S_k = \{i_1, i_2, ..., i_k\}$ corresponding to the $k$ highest scores of the scoring function $f_i(X,Y)$. 
In our experiments, we use the chi-square statistic as the score function $f_i$.

\section{Classification with Fiedler Vectors}
\label{sec:Fiedler}
In this section we apply the Fiedler Vector Algorithm, initially introduced as a theoretical framework for spectral clustering \cite{Fiedler1973, Fiedler1975}, to groups of images from the ImageNet dataset. The algorithm is based on the bit vectors collected at various layers within the neural network known as ResNet-50 \cite{ResnetHe}. Prior to delving into the applications of the algorithm with ResNet-50 bit vectors, we first provide a review of the Fiedler method and its associated definitions.

\subsection{Fiedler Vector Algorithm}
Consider a weighted graph $G=(V,E,W)$ where $V=\{1,\ldots,n\}$ denotes the set of vertices, $E$ denotes the set of (unordered) edges, and $W:E\rightarrow \mathbb R_{\geq 0}$ denotes the weights of the elements in $E$. Thus each element of $E$ is an unordered pair of vertices and has an associated non-negative weight. The {\it weighted adjacency matrix} $A \in\mathbb{R}^{n\times n}$ of the graph $G$ is defined by the rules $A_{ij}=0$ if $\{i,j\}\notin E$ and $A_{ij}=W(\{i,j\})$ if $\{i,j\}\in E$. The {\it degree matrix}, $D$, is a diagonal matrix with $D_{i,i}$ equal to the sum of the entries in the $i^{th}$ row of $A$. It can also be expressed as $D=\text{diag}(Ae)$ and $e$ is a column vector of all 1's. The Laplacian matrix of the graph $G$ can then be expressed as follows:
\begin{definition}
Given an undirected weighted graph $G=(V,E,W)$, its Laplacian matrix $L$ is defined as $L=D-A$ where $A$ is the weighted adjacency matrix of $G$ and $D$ is the degree matrix of $G$.
\end{definition}

In \cite{Fiedler1973}, it was shown that the Laplacian matrix $L$ is symmetric, positive semi-definite, and has $0$ as an eigenvalue. The multiplicity of  $0$ as an eigenvalue is equal to the number of connected components of $G$. Thus, if $G$ is a connected graph, then $0$ is an eigenvalue with multiplicity $1$. For connected graphs, the second smallest eigenvalue of $L$ is called the {\it algebraic connectivity} of the graph $G$. It provides information about the ease with which the graph can be separated into two components. In \cite{Fiedler1975}, the eigenvector associated with the second smallest eigenvalue of $L$, now referred to as the Fiedler vector, was proposed as a means of ``optimally" breaking apart $G$ into two more tightly connected components. The following theorem leads to the Fiedler Vector Algorithm for partitioning the vertices of $G$.

\begin{theorem}
Let $G=(V,E,W)$ be a weighted connected graph with weighted adjacency matrix $A$ and Laplacian matrix $L$. Let $v_2$ be an eigenvector corresponding to the second smallest eigenvalue of $L$. The Fiedler vector partition is:
\[C_1 =\{i\in N: v_2(i)<0 \}\ \text{and} \ C_2 =\{i\in N: v_2(i)>0 \}.\]
The vertices $j$ satisfying $v_2(j)=0$ can be arbitrarily included in either class.
\end{theorem}
{The aforementioned Fiedler Vector Algorithm can be extended to partition $2^l(l=2,3,\ldots)$ clusters by leveraging the sign patterns of entries in the $l$ eigenvectors corresponding to the first $l$ smallest nonzero eigenvalues.}
\subsection{Experiment Setting}
In the following sections, we will apply the Fiedler Vector algorithm to perform classification tasks on the ImageNet-1K dataset. This dataset has 1000 classes of images, each consisting of 1300 training images and 50 validation images.
ResNet-50 refers to a particular trained, deep, feed forward, convolutional, ReLU neural network that was trained on the 1.3M training images in the ImageNet-1K dataset \cite{ResnetHe}. To obtain the bit vectors for an image, we pass the image through ResNet-50 and extract the bit vector based on the output of each ReLU layer. This results in a total of 17 bit vectors (as there are 17 layers of ResNet-50 that utilize ReLU activation function).
The neural network architecture features an initial set of four ReLU layers with 802,816 nodes, followed by four layers with half the nodes, six layers with 1/4 of the nodes, and a final three layers with 1/8 of the initial number of nodes.

The weighted adjacency matrix $A$ for a given set of images at each layer is obtained by computing the matrix $\mathbf{1} - H$, where $\mathbf{1}$ is a matrix of all ones, and $H$ is the normalized Hamming distance matrix. In other words, at a given layer, the entry 
$H_{i,j}$ of $H$ is defined as the number of different entries in the bit vectors for images $i$ and $j$ divided by the length of the bit vectors.

To account for the large number of nodes in each layer and the varying degrees of importance among them, we utilized a feature selection technique described in Section~\ref{sec:featureselection} to identify the most crucial bits in the bit vector. We then labeled each image with a much shorter bit vector consisting of the most crucial bits. Finally, we applied the Fiedler Vector algorithm to the weighted adjacency matrix described in the preceding paragraph.

In the classification process, we used the training data of the selected classes to determine the $1000$ most essential features (bits), which we then used to compute the Laplacian matrices for the training and test data of the chosen classes, respectively.  
Finally, we evaluated the classification rate using the Fiedler Vector algorithm on both the training and test datasets.
\subsection{Two-Class Classification}
We defined four pairs of distinct classes for our experiments. The first two pairs 
each consists of two different species, specifically Tench vs Thunder Snake, and Rhodesian Ridgeback vs Monarch Butterfly. The other two pairs belong to subcategories of two main categories, fish and snake. In particular, the third pair is Tench vs Goldfish, and the fourth pair is Thunder Snake vs Ringneck Snake.

\begin{figure*}
 \captionsetup{skip=-3pt}
  \centering
  \includegraphics[width=0.98\linewidth]{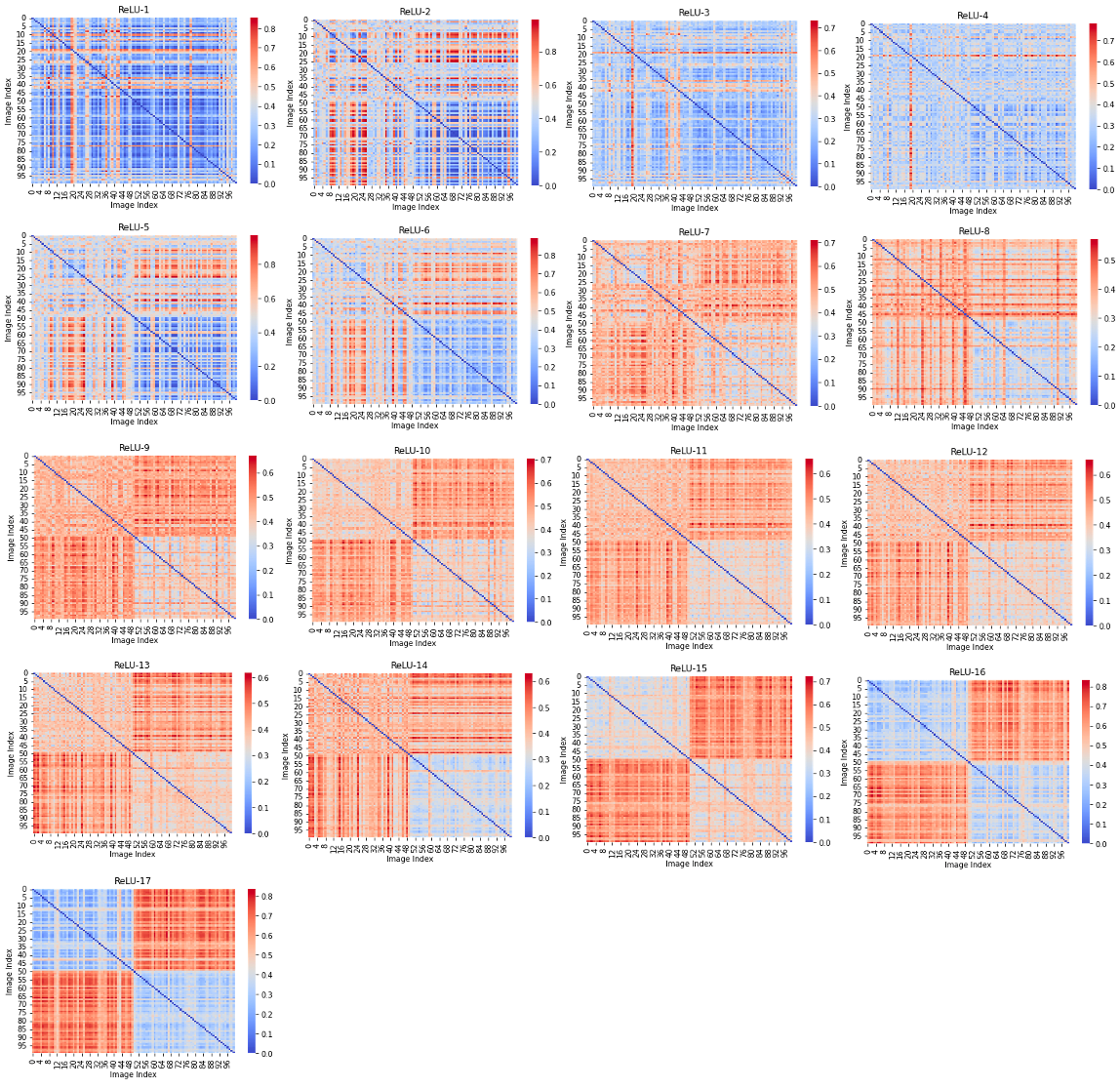}
  \caption{RDM for Tench and Thunder Snake on the test data at Layers 1-17.}
  \label{fig:RDM_layers}
\end{figure*}

 In our initial experiment, we extracted 1000 significant features from the training data for two classes (Tench and Thunder Snake) and utilized the same features from the validation set to construct an RDM. We investigated how the RDMs of these two classes evolve throughout the 17 ReLU layers as depicted in Figure \ref{fig:RDM_layers}.
 We also noticed an interesting pattern where one class grows more similar to the other in the initial layers (layers 2-6), but later, both classes drift apart as indicated by the increasing dissimilarity values.
 We observed clustering as the pairwise distance decreased within the same class, even in the initial layer. In the later layers, both classes were distinctly separated in the RDM, with a discernible boundary between them. Interestingly, the separation was already evident at the 15th ReLU layer.

\begin{figure}[h]
\centering
    \includegraphics[width = 3.2in, height = 1.4in]{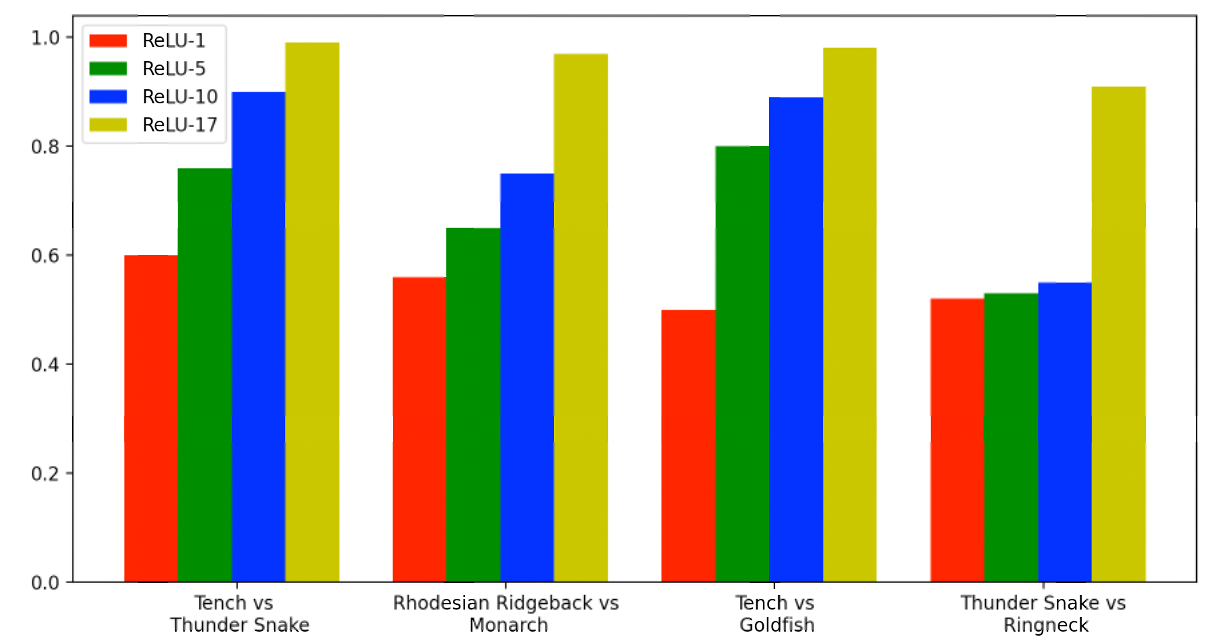}
    \caption{Test accuracy for four pairs of two classes each at layer 1, 5, 10, 17 using Fiedler Vector Algorithm.}
    \label{fig:relu_1_5_10_17}
\end{figure}

Table \ref{tab: accuracy} presents the classification accuracy for each layer and the accuracy for each class calculated using the Fiedler Vector Algorithm. The results are in line with the RDM plots. It is interesting to note that while the classification rate as a whole may decrease across two consecutive layers, the accuracy of one class may improve.
Similar experiments were performed for the remaining three pairs, and the resulting test accuracy was plotted for 1, 5, 10, and 17 ReLU layers (see Figure~\ref{fig:relu_1_5_10_17}). Our findings again indicated a positive correlation between layer depth and classification accuracy, implying that the neural network can extract increasingly valuable information from the data later in its architecture. 

Figure~\ref{fig:relu_1_5_10_17} shows comparable performance for the three pairs of classes but slightly worse performance for the fourth, Thunder Snake vs Ringneck Snake.
This can perhaps be attributed to the
fine-grained differences between the Thunder Snake and Ringneck Snake that the coarse characterization given by bit vectors may be unable to completely capture.

In conclusion, our experiments have demonstrated that the coarse representation given by bit vectors, filtered to 1000 significant features, capture enough information about the data to be used as a tool for neural network analysis. The Fiedler Vector algorithm applied to the Laplacian matrix constructed from the bit vectors is shown to be an effective method for image classification, with the classification rate generally increasing in later layers.



\begin{table*}[t]
  \centering
  \resizebox{0.8\textwidth}{!}{%
  \begin{tabular}{*{18}{c}}
    \hline
    \small{Layer number } & 1 & 2 & 3 & 4 & 5 & 6 & 7 & 8  &9 &10 &11 & 12 &13 &14&15 &16 &17\\
    \hline
      \small{Accuracy (\%)} & 60 & 69 & 58 & 52 & 76 & 81 & 83 & 59 &  84 & 90 & 83 & 85 & 78 & 73 & 100 & 100 &99 \\
         \hline
      \small{Tench (\%)} & 24 & 54 & 22& 4 & 62 & 62 & 82 & 18 & 74 &86 & 66& 72 & 56 & 46 & 100 & 100 &98  \\
         \hline
      \small{Thunder Snake (\%)} & 96 & 84 & 94 & 100 &90 & 100 & 84 & 100 & 94 & 94 & 100 & 98 &100 & 100 & 100 & 100  &100\\
      \hline
  \end{tabular}%
}
  \caption{Test accuracy (in percentage) at layer 1-17 for Tench and Thunder Snake using Fiedler Vector Classifier.}
   \label{tab: accuracy}
\end{table*}

\subsubsection{Two Superclass Classification}
\captionsetup{skip=4pt}
\begin{figure}[ht]
\centering
    \includegraphics[width = 3.3in, height = 1.4in]{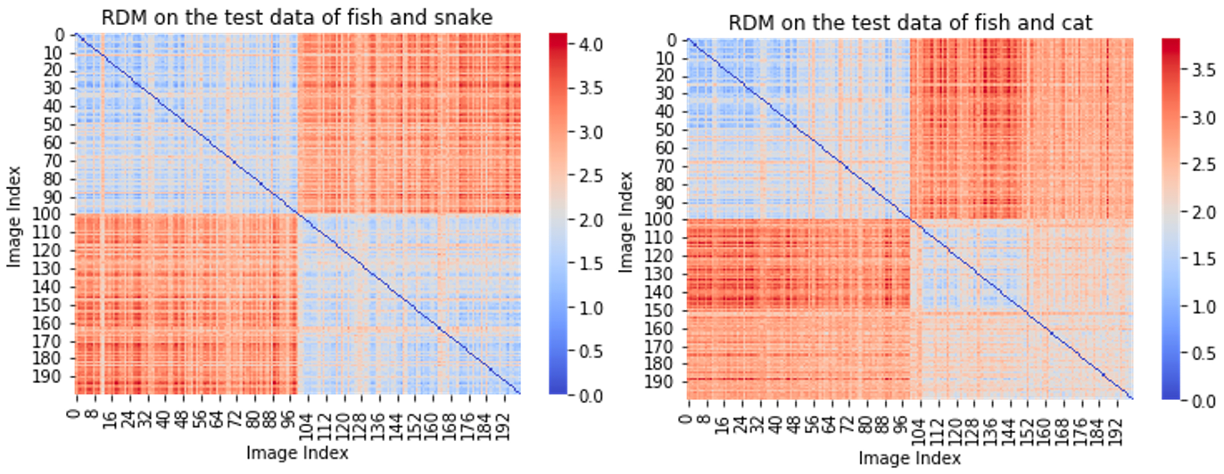}
    \caption{Test accuracy for fish (Tench and Goldfish) vs snake (Thunder and Ringneck) and for  fish  (Tench and Goldfish) vs cat (Persian and Egyptian) at layer 17 using Fiedler Vector Algorithm.}
    \label{fig:twocategories}
\end{figure}

The Fiedler vector has shown promising results in classifying two distinct classes, which prompted us to investigate its effectiveness in partitioning broader classes, specifically the superclass of fish and snake, and fish and cat. 
Our analysis indicates that the Fiedler Vector algorithm is an effective method for classifying two superclasses. To identify the best 1000 features, we selected the training datasets of class Tench and class Goldfish, and class Thunder snake and class Ringneck snake. These classes belong to the fish and snake superclasses, respectively. We then applied the Fiedler Vector algorithm to the validation dataset of these superclasses, resulting in a test accuracy of 94.5\%, 97.5\%, and 99\% for the last three ReLU layers, respectively.

We also conducted similar experiments on the fish (Tench and Goldfish) and cat (Persian and Egyptian) superclasses. For the last three ReLU layers, the resulting test accuracy was 98\%, 99\%, and 99.5\%, respectively. Figure~\ref{fig:twocategories} shows the resulting RDMs for these two pairs of superclasses at the last ReLU layer.

Our findings suggest that the bit vectors can identify common features for classes belonging to the same superclass, even in the earlier ReLU layers. Additionally, the Fiedler Vector algorithm effectively distinguishes between two different superclasses.

\section{Adversarial Image Analysis}
\label{sec:adversarial}

The success of Fiedler partitioning in separating image classes motivated us to see if bit vectors could also distinguish between adversarial and non-adversarial images.
In the next two sections, we conducted experiments to analyze different adversarial attacks. Firstly, we examined the correlation between the RDMs of bit vectors and the RDMs of latent layer embeddings. Encouraged by the positive results of this experiment, we constructed a linear SVM classifier in the subsequent section. The SVM classifier was trained on the bit vectors and achieved promising results, demonstrating the potential of using bit vectors for adversarial image detection.
\subsection{RDM Comparison between ReLU layer and Latent Layer Embeddings}
\begin{figure}[h]
\captionsetup{skip=4pt}
\centering
    \includegraphics[width = 3.4in, height = 1.4in]{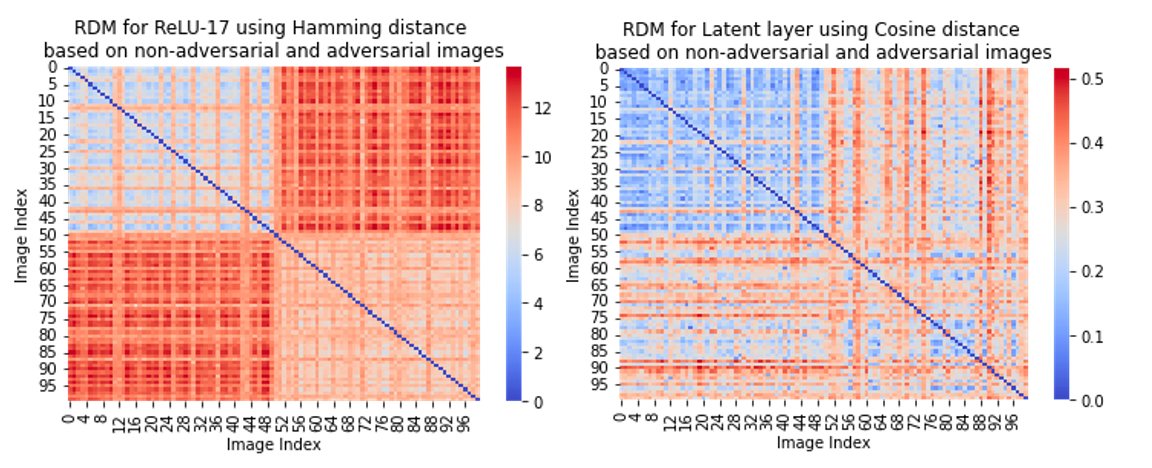}
    \caption{RDM comparisons of ReLU layer and latent layer for non adversarial and adversarial images.}
    \label{fig:motivation_svm}

\end{figure}

In our investigation, we aimed to compare the differences in latent layer embeddings and corresponding bit vectors of adversarial and non-adversarial images in the Imagenet-1K validation set. To ensure coherence, we used 2048 significant features for the ReLU-17 layer bit vectors, corresponding to the dimensionality of the latent layer embeddings. We used Hamming distance for the bit vector RDM and cosine distance for the latent layer RDM to build our RDMs. Figure \ref{fig:motivation_svm} illustrates a distinct separation between the two classes in the latent layer RDM, as anticipated. Similarly, the ReLU layer RDM showed a clear difference between adversarial and non-adversarial classes. The Pearson correlation between the two RDMs was 0.620, indicating a positive correlation between the representations.

In the following section, we investigate bit vectors' potential in distinguishing between these two types of images via a linear classifier.

\subsection{Linear Separability Comparison}
The ReLU activation patterns of deep neural networks are a tremendous simplification of the workings of the network yet contain essential information about the input data, as demonstrated by the findings in \cite{jamil2022dual}. That study indicated that only 5 bits out of the 6 million in the bit vectors are sufficient to  distinguish between most adversarial and non-adversarial images. To the best of our knowledge, the use of bit vectors to investigate adversarial images is not prevalent in the literature. However, \cite{Gorbett_2022_WACV} showed that linear separability can be achieved between adversarial and non-adversarial images using latent layer embeddings. In our experiments, we observed that bit vectors can also achieve comparable or better results in linear separability. Specifically, we utilized the last ReLU layer of ResNet-50, ReLU-17, to generate the 100352-dimensional bit vectors.

Although the linear separability using bit vector representations from the later part of the network may seem obvious, our approach's efficiency lies in selecting 1000 of the 100352 bits that contain the most information. By training a linear SVM on this concentrated information, we were able to distinguish adversarial from non-adversarial images. 
In our experimental analysis, we utilized the validation dataset of ImageNet-1K. Instead of using the entire dataset, we 
selectively choose classes based on their placement within the WordNet hierarchy \cite{miller1998wordnet}. These datasets are as follows:

\paragraph{Randomly selected classes}
We utilized a dataset consisting of 100 classes that were randomly selected from the validation dataset of ImageNet-1K. Each class contained 50 images, resulting in a total of 5000 images for the dataset.

\paragraph{Broad Classes Using WordNet Hierarchy}
The second dataset was generated by leveraging the WordNet hierarchy to identify and select high-level classes from the ImageNet validation dataset. The data was labeled based on the top-level categories, resulting in a dataset of 11,600 images that span five broad classes: 
aquatic animals, reptiles, carnivores, insects, and natural objects.

\paragraph{Subclasses from Selected Broad Classes}
The third dataset consisted of one subclass from each of the five classes in the second dataset. The selected subclasses were 
fish, snakes, working dogs, butterflies, and plants, resulting in a total of 2,650 images.


\

To generate adversarial images, we utilized four distinct attack methods: DAmageNet\cite{9238430} (the dataset was directly downloaded), Fast Gradient Sign Method (FGSM)\cite{goodfellow2014explaining}, Projected Gradient Descent (PGD)\cite{madry2017towards}, and Carlini \& Wagner (CW)\cite{7958570}.
We then conducted five Linear SVM experiments on the resulting datasets, including ImageNet vs. DAmageNet, ImageNet vs. FGSM, ImageNet vs. PGD, ImageNet vs. CarliniWagner, and ImageNet vs. all adversarial images combined.
\begin{table*}[]
\captionsetup{skip=-4pt} 
\footnotesize
\begin{center}
\begin{tabular}{|c|ccc|ccc|ccc|}
\hline
\multirow{2}{*}{Attack type} & \multicolumn{3}{c|}{Random selection}                                                                                                                                                                                                          & \multicolumn{3}{c|}{WordNet hierarchy}                                                                                                                                                                                                         & \multicolumn{3}{c|}{Subclasses}                                                                                                                                                                                                                \\ \cline{2-10} 
                             & \multicolumn{1}{c|}{\begin{tabular}[c]{@{}c@{}} $\text{ReLU}_{1000}$\\ accuracy \end{tabular}} & \multicolumn{1}{c|}{\begin{tabular}[c]{@{}c@{}}$\text{ReLU}_{2048}$\\  accuracy\end{tabular}} & \begin{tabular}[c]{@{}c@{}}Latent layer \\ accuracy\end{tabular} & \multicolumn{1}{c|}{\begin{tabular}[c]{@{}c@{}}$\text{ReLU}_{1000}$ \\ accuracy\end{tabular}} & \multicolumn{1}{c|}{\begin{tabular}[c]{@{}c@{}}$\text{ReLU}_{2048}$\\  accuracy\end{tabular}} & \begin{tabular}[c]{@{}c@{}}Latent layer \\ accuracy\end{tabular} & \multicolumn{1}{c|}{\begin{tabular}[c]{@{}c@{}}$\text{ReLU}_{1000}$ \\ accuracy\end{tabular}} & \multicolumn{1}{c|}{\begin{tabular}[c]{@{}c@{}}$\text{ReLU}_{2048}$\\  accuracy\end{tabular}} & \begin{tabular}[c]{@{}c@{}}Latent layer \\ accuracy\end{tabular} \\ \hline
DAmagenet                    & \multicolumn{1}{c|}{{0.978}}                                                  & \multicolumn{1}{c|}{{0.992}}                                                  & 0.977                                                            & \multicolumn{1}{c|}{0.961}                                                           & \multicolumn{1}{c|}{{0.989}}                                                  & 0.984                                                            & \multicolumn{1}{c|}{0.944}                                                           & \multicolumn{1}{c|}{{0.989}}                                                  & 0.968                                                            \\ \hline
FGSM                         & \multicolumn{1}{c|}{0.935}                                                          & \multicolumn{1}{c|}{{0.981}}                                                  & 0.9625                                                           & \multicolumn{1}{c|}{{0.966}}                                                  & \multicolumn{1}{c|}{{0.993}}                                                  & 0.954                                                            & \multicolumn{1}{c|}{{0.966}}                                                  & \multicolumn{1}{c|}{{0.988}}                                                  & 0.938                                                            \\ \hline
PGD                          & \multicolumn{1}{c|}{{0.917}}                                                  & \multicolumn{1}{c|}{{0.97}}                                                   & 0.902                                                            & \multicolumn{1}{c|}{{0.961}}                                                  & \multicolumn{1}{c|}{{0.979}}                                                  & 0.885                                                            & \multicolumn{1}{c|}{{0.911}}                                                  & \multicolumn{1}{c|}{{0.977}}                                                  & 0.868                                                            \\ \hline
CarliniWagner                & \multicolumn{1}{c|}{0.958}                                                           & \multicolumn{1}{c|}{{0.992}}                                                  & 0.962                                                            & \multicolumn{1}{c|}{0.959}                                                           & \multicolumn{1}{c|}{0.991}                                                           & 0.996                                                            & \multicolumn{1}{c|}{0.970}                                                           & \multicolumn{1}{c|}{0.961}                                                           & 0.993                                                            \\ \hline
All                          & \multicolumn{1}{c|}{{0.959}}                                                 & \multicolumn{1}{c|}{{0.979}}                                                  & 0.9416                                                           & \multicolumn{1}{c|}{{0.975}}                                                  & \multicolumn{1}{c|}{{0.987}}                                                  & 0.939                                                            & \multicolumn{1}{c|}{{0.936}}                                                  & \multicolumn{1}{c|}{{0.986}}                                                  & 0.931                                                            \\ \hline
\end{tabular}
\end{center}
\caption{Accuracy comparison between bit vectors and latent layer embeddings.}
\label{tab:acc}
\end{table*}

\begin{table*}[]
\captionsetup{skip=-4pt} 
\footnotesize
\begin{center}
\begin{tabular}{|c|ccc|ccc|ccc|}
\hline
                              & \multicolumn{3}{c|}{Random selection}                                                                                                                                                                                     & \multicolumn{3}{c|}{WordNet hierarchy}                                                                                                                                                                                    & \multicolumn{3}{c|}{Subclasses}                                                                                                                                                                                           \\ \cline{2-10} 
\multirow{-2}{*}{Attack type} & \multicolumn{1}{c|}{\begin{tabular}[c]{@{}c@{}}$\text{ReLU}_{1000}$ \\  AUROC\end{tabular}} & \multicolumn{1}{c|}{\begin{tabular}[c]{@{}c@{}}   $\text{ReLU}_{2048}$\\ AUROC\end{tabular}} & \begin{tabular}[c]{@{}c@{}}Latent layer\\  AUROC\end{tabular} & \multicolumn{1}{c|}{\begin{tabular}[c]{@{}c@{}}$\text{ReLU}_{1000}$ \\  AUROC\end{tabular}} & \multicolumn{1}{c|}{\begin{tabular}[c]{@{}c@{}}$\text{ReLU}_{2048}$ \\  AUROC\end{tabular}} & \begin{tabular}[c]{@{}c@{}}Latent layer\\  AUROC\end{tabular} & \multicolumn{1}{c|}{\begin{tabular}[c]{@{}c@{}}$\text{ReLU}_{1000}$ \\  AUROC\end{tabular}} & \multicolumn{1}{c|}{\begin{tabular}[c]{@{}c@{}}$\text{ReLU}_{2048}$ \\  AUROC\end{tabular}} & \begin{tabular}[c]{@{}c@{}}Latent layer\\  AUROC\end{tabular} \\ \hline
DAmagenet                     & \multicolumn{1}{c|}{{0.997}}                 & \multicolumn{1}{c|}{{0.999}}                 & 0.997                                                         & \multicolumn{1}{c|}{0.993}                                                  & \multicolumn{1}{c|}{{0.999}}                                         & 0.998                                                         & \multicolumn{1}{c|}{0.989}                                                  & \multicolumn{1}{c|}{{0.999}}                                         & 0.995                                                         \\ \hline
FGSM                          & \multicolumn{1}{c|}{0.982}                                                  & \multicolumn{1}{c|}{{0.997}}                                         & 0.992                                                         & \multicolumn{1}{c|}{{0.995}}                                         & \multicolumn{1}{c|}{{0.999}}                                         & 0.988                                                         & \multicolumn{1}{c|}{{0.995}}                                         & \multicolumn{1}{c|}{{0.999}}                                         & 0.982                                                         \\ \hline
PGD                           & \multicolumn{1}{c|}{{0.976}}                                         & \multicolumn{1}{c|}{{0.996}}                                         & 0.962                                                         & \multicolumn{1}{c|}{{0.994}}                                         & \multicolumn{1}{c|}{{0.998}}                                         & 0.948                                                         & \multicolumn{1}{c|}{{0.971}}                                         & \multicolumn{1}{c|}{{0.998}}                                         & 0.936                                                         \\ \hline
CarliniWagner                 & \multicolumn{1}{c|}{0.993}                                                  & \multicolumn{1}{c|}{{0.999}}                                         & 0.999                                                         & \multicolumn{1}{c|}{0.994}                                                  & \multicolumn{1}{c|}{{0.999}}                                         & 0.999                                                         & \multicolumn{1}{c|}{0.995}                                                  & \multicolumn{1}{c|}{0.993}                                                  & 0.999                                                         \\ \hline
All                           & \multicolumn{1}{c|}{{0.985}}                 & \multicolumn{1}{c|}{{0.996}}                                         & 0.97                                  & \multicolumn{1}{c|}{{0.994}}                 & \multicolumn{1}{c|}{{0.998}}                                         & 0.963                                 & \multicolumn{1}{c|}{{0.978}}                 & \multicolumn{1}{c|}{{0.998}}                                         & 0.955                                 \\ \hline
\end{tabular}
\end{center}
\caption{AUROC comparison between bit vectors and latent layer embeddings.}
\label{tab:auc}
\end{table*}

\subsubsection{Experiments}

We began by dividing our two sets of original images' bit vectors $X^{\text{org}}$ and adversarial images' bit vectors $X^{\text{adv}}$ into separate training and test sets, which resulted in four sets of bit vectors: $X_{\text{train}}^{\text{org}}$, $X_{\text{test}}^{\text{org}}$, $X_{\text{train}}^{\text{adv}}$, and $X_{\text{test}}^{\text{adv}}$. We applied a feature selection method to $X_{\text{train}}^{\text{org}}$ and $X_{\text{train}}^{\text{adv}}$ independently to select $k$ features (bits) from the last ReLU layer for both original and adversarial images. We discarded the other ReLU features, resulting in $X_{\text{trainSelected}}^{\text{org}}$ and $X_{\text{testSelected}}^{\text{org}}$ for original bit vectors, and $X_{\text{trainSelected}}^{\text{adv}}$ and $X_{\text{testSelected}}^{\text{adv}}$ for adversarial bit vectors. We then concatenated the original bit vectors with the adversarial bit vectors, giving us our final train and test sets: $X_{\text{train}}$ and $X_{\text{test}}$. Simultaneously, we generated labels vectors $y_{\text{train}}$ and $y_{\text{test}}$, where the label is either original or adversarial. Finally, we trained a linear SVM classifier on $(X_{\text{train}}, y_{\text{train}})$ and evaluated the classifier performance on the test set $(X_{\text{test}}, y_{\text{test}})$. The pseudo-code for this experiment is presented in Algorithm~\ref{alg1}. 

Using a consistent data split, we generated and evaluated models using two $k$ values: 2048 and 1000. Results are shown in \ref{tab:acc} and \ref{tab:auc}, corresponding with accuracy and AUROC, respectively. We selected $k = 2048$ because that corresponds with the dimensionality of the latent layer embeddings, allowing us to compare bit vectors with latent embeddings. We experimented with $k = 1000$ to investigate how bit vectors compared to latent embeddings when the dimensionality of the bit vectors was substantially lower than the latent embeddings.

\begin{algorithm} 
\caption{Linear SVM Classifier} 
\label{alg1}
\begin{algorithmic} 
    \REQUIRE ${X^{\text{org}}}$, ${X^{\text{adv}}}$, ResNet-50
    \ENSURE Accuracy and AUROC scores
    \STATE 1) Split into sets $(X^i_{\text{train}})$ and   $(X^i_{\text{test}})$ where $ i = \text{org}, \text{adv}$
    \STATE 2) Determine class labels $(y^i_{\text{train}})$ based on the number of classes indicated by the chosen dataset.
    \STATE 3) Select the top $k$ features for $(X^i_{\text{train}})$. 
    \STATE 4) Select the same features for $(X^i_{\text{test}})$.
    \STATE 5) Concatenate the selected training sets $(X^i_{\text{trainSelected}})$ for $ i = \text{org, adv}$ into a single set $(X_{\text{train}})$.
    \STATE 6) Concatenate the selected test sets $(X^i_{\text{testSelected}})$ for $ i = \text{org, adv}$ into a single set $(X_{\text{test}})$.
    \STATE 7) Create label vectors $(y_{\text{train}})$ and $(y_{\text{test}})$ for the training set and test set with 0's for original images and 1's for adversarial images.
    \STATE 8) Train a Linear SVM on $(X_{\text{train}},y_{\text{train}})$.
    \STATE 9) Evaluate the performance on $(X_{\text{test}},y_{\text{test}})$.
\end{algorithmic}
\end{algorithm}

\subsubsection{Results}
We conducted a comprehensive evaluation of our approach 
 using accuracy and AUROC metrics, compared to SVM on latent layer embeddings (2048 features) in Tables \ref{tab:acc} and \ref{tab:auc}. 
Our findings indicate that bit vectors with 2048 features exhibit better accuracy and AUROC scores than latent layer embeddings  for adversarial image detection. Also, it is worth noting that when we reduced the number of features to 1000, bit vectors performance is comparable to latent layer scores, indicating that bit vectors do not need much information to be impactful. This corresponds with the findings in \cite{jamil2022dual}, who also noticed that bit vectors can be heavily down sampled while still maintaining meaningful information. This highlights the potential for a bit vector based approach for detecting adversarial attacks, as it can extract crucial information with fewer bits.


\section{Conclusion}
This paper examined the potential of ReLU activation patterns (bit vectors) for interpreting neural networks. We investigated a novel way to extend RDMs, a popular technique for comparing the similarity of networks, to facilitate Fiedler partitioning. Through the use of Representational Dissimilarity Matrices (RDMs), we investigated the coherence of input data within the network's embedding spaces. Bit vectors were utilized to construct similarity scores between images from two distinct classes within each layer of a deep neural network. Fiedler partitioning was then employed to separate the classes in these matrices. Our findings demonstrated that the bit vectors improve the detectability of classes throughout the network, with the final ReLU layer achieving over 95\% separation accuracy. Furthermore, we showed that bit vectors are effective in adversarial image detection, achieving over 95\% accuracy in separating adversarial and non-adversarial images using a linear SVM. \label{sec:conclusion}
\section{Acknowledgements}
This work is partially supported by the United States Air Force under Contract No. FA865020C1121 and the DARPA Geometries of Learning Program under Award No. HR00112290074.


{\small
\bibliographystyle{ieee_fullname}
\bibliography{egbib}
}

\end{document}